\documentclass[journal]{IEEEtai}
\usepackage[colorlinks,urlcolor=blue,linkcolor=blue,citecolor=blue]{hyperref}
\usepackage{etoolbox}
\makeatletter
\patchcmd{\@makecaption}
  {\scshape}
  {}
  {}
  {}
\makeatletter
\patchcmd{\@makecaption}
  {\\}
  {.\ }
  {}
  {}
\makeatother
\def\tablename{Table}
\usepackage{booktabs}
\usepackage{amsmath}
\usepackage{rotating}
\usepackage{color}
\usepackage{times}
\usepackage{epsfig}
\usepackage{graphicx}
\usepackage{amssymb}
\usepackage{multirow}
\usepackage{algorithm}
\usepackage{algorithmic}
\usepackage{array}
\usepackage{cite}
\usepackage{wrapfig}
\usepackage{svg}
\usepackage{titlesec}
\titlespacing\subsection{0pt}{8pt plus 2pt minus 2pt}{3pt plus 2pt minus 2pt}
\titlespacing\subsubsection{0pt}{6pt plus 2pt minus 2pt}{3pt plus 2pt minus 2pt}

\usepackage{tabularray}
\usepackage{makecell}
\usepackage{float}
\usepackage{xr}
\let\labelindent\relax
\usepackage{enumitem}
\makeatletter
\newcommand*{\addFileDependency}[1]{
  \typeout{(#1)}
  \@addtofilelist{#1}
  \IfFileExists{#1}{}{\typeout{No file #1.}}
}
\makeatother

\usepackage{diagbox}

\usepackage[utf8]{inputenc}
\usepackage[T1]{fontenc}
\usepackage{textcomp}
\usepackage{xcolor}
\usepackage{threeparttable}

\usepackage[english]{babel}
\usepackage{blindtext}
\begin{document}
\renewcommand{\arraystretch}{0.85}
\title{SeqProFT: Sequence-only Protein Property Prediction with LoRA Finetuning}
\author{Shuo Zhang, Jian K. Liu
\thanks{
This work was partially supported by BlueBEAR and Baskerville, funded by the EPSRC and UKRI through the World Class Labs scheme (EP/T022221/1) and the Digital Research Infrastructure programme (EP/W032244/1) and operated by Advanced Research Computing at the University of Birmingham.
(\textit{Corresponding author: Jian K. Liu})}
\thanks{S. Zhang and J. K. Liu are with the School of Computer Science, Centre for Human Brain Health, University of Birmingham, Edgbaston, Birmingham, B15 2TT, United Kingdom (e-mail: sxz325@student.bham.ac.uk, j.liu.22@bham.ac.uk).}
}

\maketitle

\begin{abstract}

Protein language models (PLMs) have demonstrated remarkable capabilities in learning relationships between protein sequences and functions. However, finetuning these large models requires substantial computational resources, often with suboptimal task-specific results. This study investigates how parameter-efficient finetuning via LoRA can enhance protein property prediction while significantly reducing computational demands. By applying LoRA to ESM-2 and ESM-C models of varying sizes and evaluating 10 diverse protein property prediction tasks, we demonstrate that smaller models with LoRA adaptation can match or exceed the performance of larger models without adaptation. Additionally, we integrate contact map information through a multi-head attention mechanism, improving model comprehension of structural features. Our systematic analysis reveals that LoRA finetuning enables faster convergence, better performance, and more efficient resource utilization, providing practical guidance for protein research applications in resource-constrained environments. The code is available at https://github.com/jiankliu/SeqProFT.

\end{abstract}

\begin{IEEEImpStatement}

Rapid advancements in large language models (LLMs) have transformed our ability to understand and process natural languages. Protein language models (PLMs) extend these ideas to the biological sequences of proteins, enabling a deeper understanding of their function. This study represents a major leap forward in protein language modeling by tackling the challenges of high computational demands and the lack of task-specific precision in fine-tuning existing models. The research offers a more efficient and targeted approach by applying the LoRA method to finetune the ESM-2 model for protein property prediction. Introducing a multi-head attention mechanism, which integrates sequence features with contact map information, significantly enhances the model's ability to interpret complex protein sequences. This advancement has the potential to accelerate discoveries in protein-related research using efficient LLMs.

\end{IEEEImpStatement}

\begin{IEEEkeywords}
protein property prediction, protein language model, parameter-efficient finetuning, LoRA
\end{IEEEkeywords}

\section{Introduction}
\label{sec:intro}

Protein, as a fundamental component of life, plays a vital role in living cells~\cite{alberts2022analyzing}. Understanding protein functions is important for drug discovery, personal medical treatment, and bioengineering. With the latest developments in natural language processing, protein language models (PLMs) learn to understand the relationship between protein sequences, structures, and functions by treating amino acid sequences as text~\cite{madani2023large}. Using the inherent sequential nature of protein sequences, PLMs, such as ESM-2~\cite{lin2023esm2} and ProtTrans~\cite{elnaggar2021prottrans}, learn protein evolution patterns and capture protein structure information in a self-supervised manner from a large amount of unlabeled protein sequence data. Unlike PLMs that only use protein sequences, some PLMs also use protein structure information. SaProt~\cite{su2024saprot} performs well in predicting protein function and mutation effects by using the 3D structure of the encoded protein and the residue sequences to train the model. Implicit Neural Representations (INRs)~\cite{lee2024proteininr} was proposed to combine with the protein surface to enhance the ability of PLMs to learn the representation of protein structure. 

Similar to the NLP field, there are many downstream prediction tasks regarding proteins, such as protein thermostability prediction and protein secondary structure classification. Pre-trained PLMs are usually used as feature extractors to extract high-dimensional features of protein input~\cite{chenHybridGCNProteinSolubility2023, zhouProCeSaContrastEnhancedStructureAware2025}. This process is performed offline, that is, without gradient backpropagation. The high-dimensional features are then passed as input to the designed downstream task prediction network, which is usually composed of multilayer perceptrons (MLPs), and the final predictions are obtained after training. Although the work mentioned above demonstrated the excellent performance of the proposed PLMs in downstream tasks, such a two-stage paradigm can lead to suboptimal and non-task-specific results, because the number of parameters of the downstream network is much smaller than that of the PLMs, and high-dimensional features cannot be transferred according to the domain of a certain downstream dataset~\cite{he2022towards}. Another approach is to fully finetune the PLMs for each downstream task. Obviously, this requires a lot of computing resources and is very time-consuming.

\begin{figure*}[htb]
\centering
\includegraphics[width=0.95\linewidth]{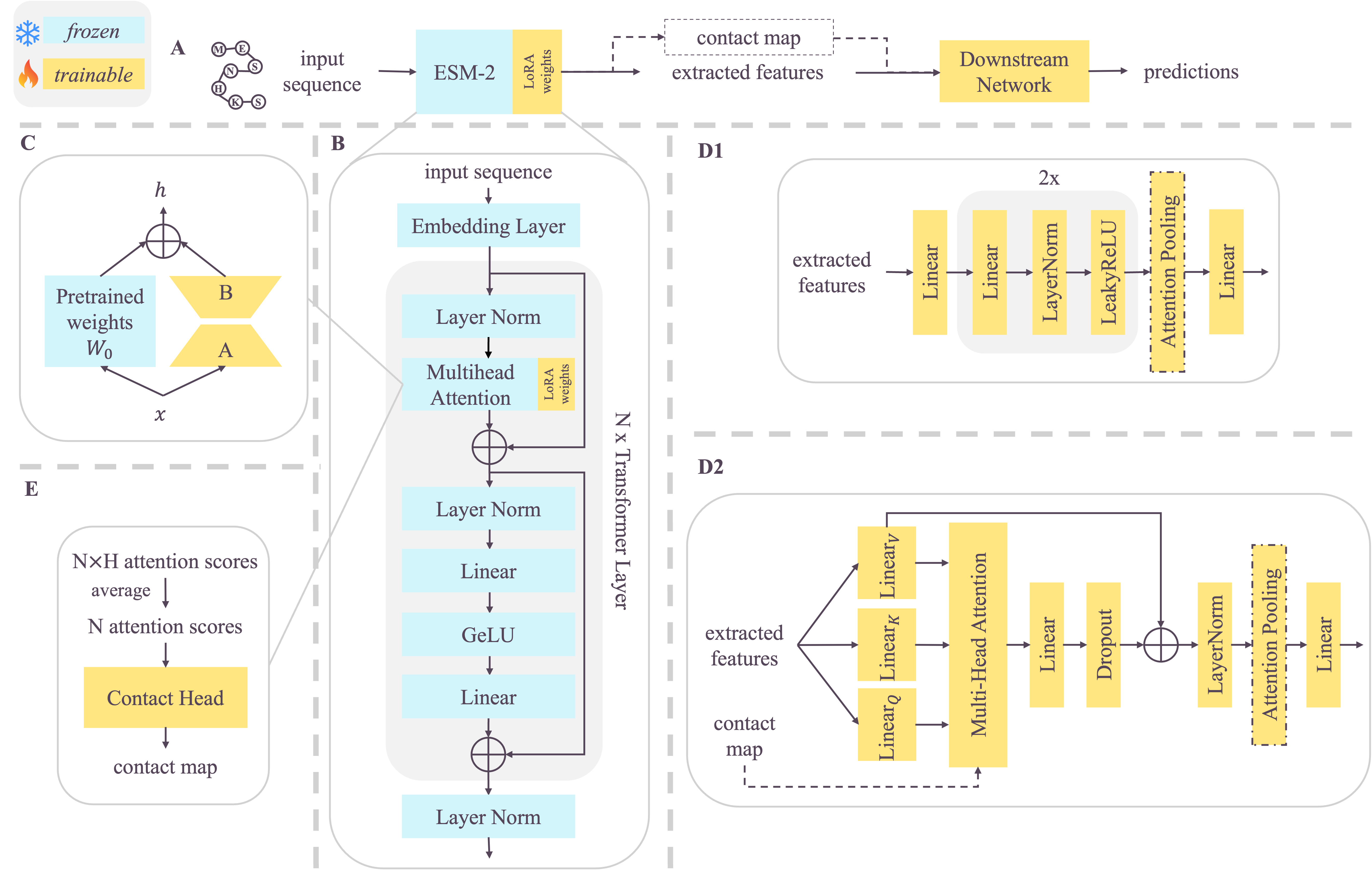}
\caption{Architecture of SeqProFT. A: Overview showing pretrained ESM-2 model with LoRA adaptation and downstream head (blue: frozen parameters; yellow: trainable parameters). B: ESM-2 structure with LoRA applied to attention layers. C: LoRA training process with frozen pretrained weights and tunable low-rank matrices. D: Downstream networks: D1 shows Simple MLP Head (SMH); D2 shows Multi-head Attention Head with optional contact map processing (MAH, CM-MAH). E: Process for generating task-specific contact maps from attention scores.}
\label{fig:arch}
\end{figure*}

To address these challenges,  the parameter-efficient finetuning (PEFT) technique~\cite{sledzieski2024demo}, specifically utilizing the LoRA (Low-Rank Adaptation)~\cite{hu2022lora} method, was introduced to improve PLMs in proteomics. This study focuses on two key tasks: predicting protein-protein interactions and determining the symmetry of homooligomer quaternary structures. It demonstrated that the PEFT approach can achieve competitive performance compared to traditional finetuning while significantly reducing memory usage and the number of required parameters. LoRA was used in ESM-2~\cite{lin2023esm2}, enabling adaptation of evolutionary knowledge embedded in protein sequences for the prediction of signal peptides~\cite{zeng2024peftsp}. Finetuning techniques were used to compare three leading PLMs (ESM-2, ProtT5~\cite{elnaggar2021prottrans}, Ankh~\cite{elnaggar2023ankh}) in eight different protein-related tasks~\cite{schmirler2023finetunetum}. In addition to LoRA, other studies have explored alternative finetuning methods. The challenges of annotating proteins with unknown functions were addressed by directly finetuning PLMs on classification tasks to significantly improve the quality of protein embeddings~\cite{dickson2024finetuning}. The finetuned embeddings not only outperform traditional classifiers in Gene Ontology (GO) annotations but also remain interpretable through similarity comparisons between proteins. Prefix-tuning of PLMs was used to generate proteins tailored for specific biomedical applications, showing that multiple properties such as antimicrobial function and alpha-helix structure can be combined into a single protein design~\cite{luo2023prefixprot}.

Due to the limitations of experimental techniques such as X-ray crystallography and cryo-electron microscopy, protein structures are more difficult to determine experimentally than sequences~\cite{chua2022cryo}. As a result, only hundreds of thousands of experimentally determined protein structures exist, whereas protein sequence data is in the billions. Most protein functional property prediction tasks only provide sequence data. Although 3D structures can be predicted using folding models such as ESMFold~\cite{lin2023esm2}, the introduced bias and time consumption cannot be ignored. Any errors in predicted structures propagate to downstream tasks, compounding uncertainty, especially for proteins with low-confidence structural predictions. Therefore, finetuning using only protein sequences is a more time-efficient method and has been shown to achieve performance comparable to methods that combine structural data and other modalities~\cite{lin2023esm2}. Sequence-based finetuning can not only alleviate the data scarcity problem but also exploit the rich information encoded in the sequence itself, making it a practical alternative for many tasks. 

This work aims to systematically investigate parameter-efficient finetuning for protein property prediction using only sequence information as input. While previous studies have applied such techniques to specific protein tasks, our work provides a comprehensive evaluation across 10 diverse prediction tasks using ESM-2 and ESM-C models of different sizes. We demonstrate that smaller models with LoRA finetuning can achieve comparable or superior performance to larger models without finetuning, offering significant practical advantages for resource-constrained research environments. To enhance model performance, we integrate the multi-head attention mechanism in the downstream head to combine sequence features with contact maps, enabling the model to consider both sequential and structural relationships during prediction. This approach involves training the LoRA weights and downstream prediction head while also allowing the model contact head to participate in training when structural information is utilized. Our experimental results confirm that our method performs well in both accuracy and computational efficiency across various parameters and tasks, providing an effective solution for protein sequence-based regression and classification problems.

\section{Methods}
\label{sec:methods}

Figure~\ref{fig:arch}.A illustrates the workflow for predicting properties of protein sequences using a pretrained ESM-2 and LoRA weights. The frozen (blue) components are kept fixed during training, while the trainable components (yellow, like the LoRA weights) are adjustable. The ESM-2 model processes the input sequences and extracts features, which may include a contact map according to the downstream network. These features are passed along to a downstream head, which further processes them to generate task-specific predictions. 

\subsection{Protein Language Model ESM}
ESM models are a series of PLMs designed for protein tasks. In this work, we evaluate ESM-2~\cite{lin2023esm2} and ESM-C~\cite{esmteam2024esmc}, where both share a similar model design. ESM-2 is a large-scale transformer-based language model designed to predict protein structures directly from amino acid sequences (Figure~\ref{fig:arch}.B). The core of the model is trained using a masked language modeling objective, where it predicts the identity of masked amino acids in protein sequences based on the surrounding context. ESM-2 scales from 8 million to 15 billion parameters. As the model's size increases, it captures increasingly complex patterns in protein sequences, which correspond to their underlying biological structures. 

The contact head in ESM-2 plays a crucial role in predicting residue-residue contact maps, which are essential for understanding the protein's 3D structure. This component extracts structural information from the attention scores generated by the transformer layer, linking sequence patterns to the physical interactions between amino acids. Specifically, for a protein sequence $X=(x_1, x_2, ..., x_L)$ of length $L$, the attention mechanism generates attention score $C_{ij}\in \mathbb{R}^{1\times 1}$ that indicates the relevance of residue $x_j$ to residue $x_i$:
\begin{equation}
    C_{ij}=softmax(\frac{Q_iK^T_j}{\sqrt{d_k}}),
\end{equation}
where $Q_i\in \mathbb{R}^{1\times d_k}$ and $K_j\in \mathbb{R}^{1\times d_k}$ are the query and key vectors for residue $x_i$ and $x_j$, respectively, and $d_k$ is the dimension of the key vectors. After considering the interactions of all residues, we obtain the attention scores $C\in \mathbb{R}^{L\times L}$ for sequence X. 

During training, multiple layers and attention heads are used in the transformer model. In the original ESM-2 setting, the attention scores from all layers and attention heads are aggregated by stacking them and then averaging them. Considering that the stacking operation will consume a large amount of memory when introducing the batch dimension, we choose to first average the attention scores across the attention head dimension and then proceed with layer-wise stacking (Figure~\ref{fig:arch}.E). The specific calculation equation is as follows:
\begin{equation}
    \tilde{C}=\frac{1}{H_1}\sum_{h=1}^{H_1}C_h,
\end{equation}
where $H_1$ is the number of attention heads, $C_h$ is the attention scores of the $h$-th attention head, and $\tilde{C}\in \mathbb{R}^{L\times L}$. In biological terms, if residue $i$ is in contact with residue $j$, the residue $j$ must also be in contact with residue $i$. Therefore, the attention score matrix must be symmetric, i.e., $\tilde{C}_{ij} = \tilde{C}_{ji}$, which is achieved by:
\begin{equation}
    \tilde{C}^{symm}=\tilde{C}+\tilde{C}^T.
\end{equation}

Afterwards, we stack the symmetric attention scores from all layers:
\begin{equation}
    \hat{C}=[\tilde{C}^{symm}_{1},\tilde{C}^{symm}_{2},\cdots ,\tilde{C}^{symm}_{N}],
\end{equation}
where $N$ is the number of transformer layers and $\hat{C}\in \mathbb{R}^{N\times L\times L}$.  

To convert these attention scores into contact probabilities $P$ (i.e., the likelihood that two residues are in contact), ESM-2 applies a linear projection that is given by:
\begin{equation}
    P=\sigma(W_p\cdot \hat{C}+b),
\end{equation}
where $W_p\in \mathbb{R}^{N\times 1}$ and $b\in \mathbb{R}$ are learnable weights and bias, and $\sigma(\cdot)$ is the sigmoid activation function, ensuring that the output contact map $P\in \mathbb{R}^{L\times L}$ consists of probabilities between 0 and 1.

\subsection{Low-Rank Adaptation (LoRA)}
LoRA injects the trainable low-rank decomposition matrices into transformer models, while freezing the pretrained weights. This approach significantly reduces the number of trainable parameters required for downstream tasks. Specifically, given the pretrained weights $W_0\in \mathbb{R}^{d\times k}$, the accumulated gradient update $\Delta W$ is represented by two trainable low-rank decomposition matrices $A\in \mathbb{R}^{r\times k}$ and $B\in \mathbb{R}^{d\times r}$:
\begin{equation}
    W_0+\Delta W=W_0+BA
\end{equation},
where $r\ll \min(d,k)$ is the rank of a LoRA module. $A$ and $B$ are initialized with a random Gaussian distribution and zero, respectively. As shown in Figure~\ref{fig:arch}.C, the forward propagation $h=W_0x$ after inserting LoRA is given by:
\begin{equation}
    h=W_0x+\Delta W x=W_0x+BAx.
\end{equation}
In addition, $\Delta Wx$ is scaled by $\frac{\alpha}{r}$ for stable training, where $\alpha$ is a constant hyperparameter. Unless otherwise specified, the rank $r$ and scale factor $\alpha$ are set to 32 in this work.

\subsection{Downstream Networks}

The ESM-2 extracted high-dimensional features are passed into downstream networks for task-specific finetuning. All parameters in downstream nets are trainable. We design 3 types of downstream networks.

\subsubsection{Simple MLP Head}
The simple MLP head (SMH) shown in Figure~\ref{fig:arch}.D1 consists of one linear layer, 2 feedforward blocks, one attention pooling layer, and the final prediction layer. Similar to~\cite{chen2021structure}, the attention pooling layer aggregates residue-level features into protein-level features. Given the hidden feature $X\in \mathbb{R}^{L\times d_k}$, we first calculate the attention score $S$ by:
\begin{equation}
    S=Dropout(Softmax(\frac{W_Q(W_KX^T)^T}{\sqrt{d_l}}))\in \mathbb{R}^{H_2\times L},
\end{equation}
where $d_l$ is the latent dimension and $H_2$ is the number of attention heads. $W_Q\in \mathbb{R}^{H_2\times d_l}$ and $W_K\in \mathbb{R}^{H_2\times d_l\times d_k}$ are learnable weights. Then the hidden feature $X$ is weighted and aggregated over all the residues in the sequence by attention score $S$ to get $X_O\in \mathbb{R}^{H_2\times d_l}$, which is given by equation~\ref{eq:SWVX}, where $W_V\in \mathbb{R}^{H_2\times d_l\times d_k}$. Finally, as shown in equation~\ref{eq:WOXO}, the pooled hidden feature $X_P\in \mathbb{R}^{d_k}$ is calculated by matrix-multiplying $X_O$ with $W_O\in \mathbb{R}^{d_k\times H_2\times d_l}$ to aggregate features from all attention heads.
\begin{equation}
\label{eq:SWVX}
    X_O=S(W_VX^T)^T
\end{equation}
\begin{equation}
\label{eq:WOXO}
    X_P=W_O(X_O)^T
\end{equation}
\subsubsection{Multi-head Attention Head}
The prediction of protein properties is a complex task that requires consideration of the interactions and associations between residues. To capture the implicit relationship and improve the prediction accuracy, we utilize the multi-head attention mechanism~\cite{vaswani2017attention} to obtain the attention distributions in different sub-spaces of the input protein sequence. Our Multi-head Attention Head (MAH) (Figure~\ref{fig:arch}.D2) processes the extracted features through three linear projection layers to generate query, key, and value representations. These are fed into a multi-head attention layer with 4 attention heads, followed by a dropout layer (rate=0.1), residual connection, and layer normalization as in the standard transformer architecture~\cite{vaswani2017attention}. The processed features are then passed through the attention pooling layer and a final linear projection layer for property prediction. This structure enables the model to capture complex relationships between amino acid residues through self-attention mechanisms.

\subsubsection{Contact Map Enhanced Multi-head Attention Head}

The protein structure is a determining factor of the protein property. In our Contact Map enhanced Multi-head Attention Head (CM-MAH), we introduce the contact map as attention weights to represent the spatial relationship between residues. Unlike in SMH and MAH where the contact head of ESM-2 remains frozen, in CM-MAH we train the contact head to learn task-specific contact patterns. Therefore, the attention mechanism is given by:
\begin{equation}
\label{eq:cm-attn}
    Attention(Q,K,V,P)=softmax(\frac{QK^TP}{\sqrt{d_{head}}})V,
\end{equation}
where $d_{head}$ is the feature dimension of the downstream head. In equation~\ref{eq:cm-attn}, the original attention weight is further weighted by the learned contact map to allow the model to focus more on residue pairs that have physical contact. The other layers of CM-MAH is the same as MAH, as shown in Figure~\ref{fig:arch}.D2.

\section{Experimental Design}
\label{sec:setup}

\textbf{Downstream Tasks}: We evaluate our models across 10 diverse protein property prediction tasks spanning classification and regression problems. Each task is designed to assess the model's performance across different biological functions and structural attributes, thus providing a thorough evaluation of protein characteristics.

\textbf{Enzyme Commission number prediction (EC)}~\cite{zhang2023gearnet}, involves predicting the Enzyme Commission numbers that classify enzymes based on the chemical reactions they catalyze. Each protein sequence can belong to multiple classes out of a total of 538. Accurate EC number predictions are essential for understanding enzyme functions and their roles in metabolic pathways.

\textbf{Gene Ontology term prediction (GO)}~\cite{zhang2023gearnet}, focuses on predicting Gene Ontology terms, which provide a standardized representation of gene and gene product attributes across species. This classification task includes three subsets: Biological Process (BP) with 1,943 classes, Cellular Component (CC) with 320 classes, and Molecular Function (MF) with 489 classes. Predicting GO terms is vital for annotating protein functions and understanding their biological roles.

\textbf{Fold classification (Fold)}~\cite{hou2017fold}, aims to classify protein sequences into one of 1,195 known folds, which is critical for inferring evolutionary relationships and understanding how protein sequences relate to their structures. This dataset contains three test sets: family, fold, and superfamily.

\textbf{Secondary Structure prediction (SS)}~\cite{klausen2019ssp}, involves predicting the 3-class secondary structure of each residue in a protein chain. This task is fundamental for protein structure determination and provides insights into protein folding and stability. We use CASP12 as the test set.

\textbf{Sub-cellular Localization prediction (Loc)}~\cite{elnaggar2023ankh}, predicts the sub-cellular localization of proteins into 10 different classes, which is essential for understanding their functional context within a cell. Accurate localization predictions help clarify protein functions and interactions in various cellular compartments.

\textbf{Thermostability prediction (HumanCell)}~\cite{dallago2021flip}, is a regression task that predicts the thermostability of each protein sequence. Understanding protein thermostability is important for biotechnology applications and for designing proteins with enhanced stability for therapeutic uses. Due to limited computing resources, we choose the smallest subset, HumanCell.

\textbf{GB1 Fitness prediction (GB1)}~\cite{dallago2021flip}, is a regression task where the model learns the epistatic interactions between mutations. This is vital for engineering proteins with desired properties.

\textbf{Fluorescence Landscape prediction (FLU)}~\cite{rao2019tape}, is another regression task. The dataset is partitioned based on the Hamming distances between the parent green fluorescent protein (GFP) and its mutants. The training set includes mutants with a Hamming distance of 3 from the parent GFP, while the test set comprises more distant variants with four or more mutations. Predicting this landscape helps identify mutations that enhance or impair protein performance.

For EC, GO, and Fold, we use the dataset splits from GearNet~\cite{zhang2023gearnet}. For SS, Loc, and FLU, we follow the dataset splits provided in Ankh~\cite{elnaggar2023ankh}. For HumanCell and GB1, we adopt the dataset splits from FLIP~\cite{dallago2021flip}.

\textbf{Dataset Sequence Characteristics:} The benchmark datasets have a large variation in sequence similarity between the training and test sets (Table S1). The Loc and HumanCell datasets present a challenge for generalization due to lower sequence identity (<40\%), while GB1 and FLU have highly similar sequences (>98\%). This diversity in the data allows for a thorough evaluation of the model, as the performance improvement on the low-identity datasets could suggest a better understanding of protein function, not simply sequence recall. While high sequence similarity in GB1 and FLU could simplify the prediction task, this condition affects all models equally, enabling a direct attribution of the reported performance gains to our proposed method. Other detailed characteristics are provided in the Supplementary Section S1 and Table S1.

\textbf{Training Setup}: We evaluated three ESM-2 model sizes (35M, 150M, 650M) and two ESM-C model sizes (300M, 600M) with three downstream heads (SMH, MAH, CM-MAH) on a single A100 40G GPU. LoRA parameters are set to rank r=32 and scaling factor $\alpha$=32 unless otherwise specified. We use task-appropriate loss functions (multi-label BCE for EC/GO, cross-entropy for classification, MSE for regression) and standard evaluation metrics (F1-max, accuracy, Spearman's $\rho$ ). Loss function equations, complete hyperparameter details, and training procedures are in Supplementary Section S2, S3, and S4. Note that benchmark datasets follow the same splits as prior work, where some datasets did not provide validation sets (SS and Loc). Training epochs were determined by monitoring test set performance on these datasets, which could overestimate performance due to data leakage. However, since this procedure was applied identically to all models, the comparison between them remains valid. We therefore focus on the relative performance gains from fine-tuning, rather than the absolute performance values. Future work should implement proper validation splits for unbiased evaluation.

\section{Results}
\label{sec:resultsanddiscussion}

\subsection{LoRA enables stronger adaptation on downstream tasks}

\begin{table*}
\begin{center}
\scriptsize
\caption{Performance comparison of ESM-2 and ESM-C models with and without LoRA finetuning across 10 protein property prediction tasks. Results show performance metrics for models of varying sizes (35M-650M parameters) using CM-MAH downstream head. 'w/o': without LoRA finetuning; 'w/': with LoRA finetuning; $\uparrow$: percentage improvement with LoRA. Values represent the mean±SD (standard deviation) across different replicates with varying random seeds. Bold text indicates the best overall
performance.}
\label{tab:mha-c_loraVSnolora}
\begin{tabular}[0.9\linewidth]{cccccccccccccc}
\toprule
\multirow{2}{*}{Size} & \multirow{2}{*}{Settings}  & \multirow{2}{*}{EC} & \multicolumn{3}{c}{GO} & \multicolumn{3}{c}{Fold} & \multirow{2}{*}{SS} & \multirow{2}{*}{Loc} & \multirow{2}{*}{FLU} & \multirow{2}{*}{HumanCell} & \multirow{2}{*}{GB1}  \\ \cmidrule{4-9}
 & & & BP & MF & CC & family & fold & superfamily & & & &   \\ \midrule \midrule
 \multirow{5}{*}{\makecell{ESM-2\\35M}} & \multirow{2}{*}{w/o} & 0.740 & 0.393 & 0.528 & 0.486 & 0.932 & 0.250 & 0.557 & 0.744 & 0.743 & 0.623 & 0.678 & 0.803  \\
                           &                      & ±0.006 & ±0.002 & ±0.005 & ±0.001 & ±0.001 & ±0.012 & ±0.111 & ±0.001 & ±0.005 & ±0.004 & ±0.006 & ±0.006  \\
                           & \multirow{2}{*}{w/}  & 0.793 & 0.411 & 0.573 & 0.494 & 0.953 & 0.276 & 0.596 & 0.750 & 0.756 & 0.681 & 0.689 & 0.941  \\
                           &                      & ±0.001 & ±0.000 & ±0.001 & ±0.007 & ±0.006 & ±0.009 & ±0.010 & ±0.002 & ±0.003 & ±0.001 & ±0.005 & ±0.001  \\ \cmidrule{2-14}
                          & $\uparrow$  & +7.28\% &  +4.72\% & +8.35\% & +1.58\% & +2.33\% & +10.60\% & +7.02\% & +0.75\% & +1.70\% & +9.17\% & +1.50\% & +17.14\%  \\ \midrule
 \multirow{5}{*}{\makecell{ESM-2\\150M}} & \multirow{2}{*}{w/o} & 0.810  & 0.416  &0.586  &0.496 & 0.960 & 0.286 & 0.623 & 0.788 & 0.779 & 0.629 & 0.680 & 0.910\\
                            &                      & ±0.003 & ±0.001 &±0.001  &±0.004 & ±0.003 & ±0.004 & ±0.014 & ±0.001 & ±0.003 & ±0.005 & ±0.003 & ±0.008 \\
                            & \multirow{2}{*}{w/}  & 0.854 & 0.442 & 0.634 & 0.522 & 0.975 & 0.304 & 0.683& 0.796& 0.801 & 0.680 & 0.693 & \textbf{0.957}\\
                            &                      & ±0.004 & ±0.001 & ±0.002 & ±0.003 & ±0.003 & ±0.011 & ±0.007& ±0.002 & ±0.004 & ±0.001 & ±0.005 & ±0.002 \\ \cmidrule{2-14}
                              & $\uparrow$  & +5.47\% & +6.08\%	& +8.17\%	& +5.31\%	&  +1.56\% &  +6.50\% & +9.65\% & +1.05\%	& +2.79\%	& +8.13\% & +2.00\%	&+5.29\%	\\ \midrule
 \multirow{5}{*}{\makecell{ESM-2\\650M}} & \multirow{2}{*}{w/o} &0.838 & 0.436 & 0.612 & 0.516 & 0.984 & 0.295  & 0.666 & 0.819 & 0.801 & 0.638 & 0.682 & 0.881 \\
                            &                      &±0.007 & ±0.006 & ±0.004 & ±0.007 & ±0.004 & ±0.004  & ±0.009 & ±0.001 & ±0.006 & ±0.004 & ±0.008 & ±0.016  \\
                            & \multirow{2}{*}{w/}  &0.886 & 0.463 & \textbf{0.674} & 0.534 &  0.988 & \textbf{0.339} & \textbf{0.746} & \textbf{0.825} & \textbf{0.831} & \textbf{0.682} & \textbf{0.708} & 0.954  \\
                            &                      &±0.000 & ±0.011 & ±0.002 & ±0.015 & ±0.001 & ±0.017 & ±0.006 & ±0.002 & ±0.003 & ±0.003 & ±0.003 & ±0.003  \\ \cmidrule{2-14}
                          & $\uparrow$  & +5.82\%	& +6.02\%	& +10.20\%	& +3.48\%	&  +0.48\% & +15.12\% & +11.93\% & +0.72\%	& +3.80\% & +6.88\% & +3.84\%	&  +8.26\%	 \\ \midrule \midrule
 \multirow{5}{*}{\makecell{ESM-C\\300M}} & \multirow{2}{*}{w/o} &0.824 & 0.424 & 0.603 & 0.516 & 0.970 & 0.267  & 0.551 & 0.801 & 0.770 & 0.628 & 0.681 & 0.894  \\
                            &                      &±0.002 & ±0.002 & ±0.001 & ±0.002 & ±0.003 & ±0.008  & ±0.009 & ±0.000 & ±0.008 & ±0.009 & ±0.001 & ±0.005 \\
                            & \multirow{2}{*}{w/}  &0.878 & 0.475 & 0.672 & 0.545 &  \textbf{0.989} & 0.283 & 0.677 & 0.806 & 0.818 & 0.681 & 0.697 & 0.953  \\
                            &                      &±0.002 & ±0.003 & ±0.003 & ±0.002 & ±0.001 & ±0.003 & ±0.013 & ±0.001 & ±0.005 & ±0.002 & ±0.003 & ±0.002   \\ \cmidrule{2-14}
                          & $\uparrow$  & +6.61\%	& +12.07\%	& +11.53\%	& +5.60\% & +1.92\% & +5.91\% & +22.91\% & +0.66\%	& +6.25\% & +8.48\% & +2.30\%	&  +6.57\%	 \\ \midrule
 \multirow{5}{*}{\makecell{ESM-C\\600M}} & \multirow{2}{*}{w/o} &0.830  & 0.431  & 0.610  & 0.525  & 0.967  & 0.249   & 0.542  & 0.813  & 0.784  & 0.621 & 0.682  & 0.915   \\
                                        &                      &±0.004 & ±0.002 & ±0.003 & ±0.002 & ±0.001 & ±0.002  & ±0.000 & ±0.002 & ±0.004 & ±0.004 & ±0.001 & ±0.005  \\
                            & \multirow{2}{*}{w/}  &\textbf{0.888}  & \textbf{0.482}  & 0.673  & \textbf{0.552}  & \textbf{0.989} & 0.287  & 0.693  & 0.821  & 0.813  & \textbf{0.682}  & \textbf{0.708}  & 0.956 \\
                            &                      &±0.001 & ±0.001 & ±0.002 & ±0.002 & ±0.001 & ±0.017 & ±0.019 & ±0.002 & ±0.003 & ±0.002 & ±0.003 & ±0.000 \\ \cmidrule{2-14}
                          & $\uparrow$  &  +7.00\%	& +11.83\%	& +10.28\%	& +4.99\%	&  +2.29\% & +15.41\% & +27.84\% & +1.03\%	& +3.62\% & +9.95\% & +3.69\%	&  +4.40\%	 \\                         
\bottomrule
\end{tabular}
\end{center}
\end{table*}

We evaluated the effectiveness of LoRA finetuning across 8 datasets (10 subsets), using ESM-2 models with three parameter sizes (35M, 150M, and 650M) and ESM-C models with two parameter sizes (300M, 600M) as initial weights, and CM-MAH as the downstream task head. Table~\ref{tab:mha-c_loraVSnolora} presents the performance of each model on various tasks and quantifies the performance gains ($\Delta$) after applying LoRA finetuning. Models with LoRA (denoted as 'w/') demonstrate significant improvements across all tasks and model sizes compared to those without LoRA finetuning (denoted as 'w/o'). 

For the ESM-2-35M model, LoRA finetuning consistently enhances model performance, showing substantial gains in all tasks, particularly in GB1 (+17.14\%), Fold-fold (+10.60\%), and FLU (+9.17\%). The 150M model exhibits higher baseline performance, and with LoRA finetuning, achieves notable improvement across all tasks, with the maximum improvement reaching  9.65\% on the Fold-superfamily task. The 650M model, with the largest parameter count among ESM-2 variants, delivers the best overall results, achieving the highest gains in tasks such as Fold-fold (+15.12\%), Fold-superfamily (+11.93\%), and GO-MF (+10.2\%).

The 300M ESM-C model shows particularly strong gains in Fold-superfamily (+22.91\%), GO-BP (+12.07\%), and GO-MF (+11.53\%). The 600M ESM-C model achieves similar performance patterns, with notable improvements in Fold-superfamily (+27.84\%), Fold-fold (+15.41\%), and GO-BP (+11.83\%), demonstrating the effectiveness of LoRA finetuning across different model architectures.

Interestingly, our results reveal that smaller models with LoRA finetuning can achieve comparable or even superior performance to larger models without LoRA. For instance, the ESM-2-150M model with LoRA finetuning outperforms the ESM-2-650M model without LoRA on several tasks, including EC (0.854 vs 0.838), GO-BP (0.442 vs 0.436), and GB1 (0.957 vs 0.881). This represents a remarkable 75.3\% reduction in total parameters while achieving better results, demonstrating the practical value of parameter-efficient adaptation over simply scaling model size.

The parameter efficiency gains are substantial across all model sizes as shown in Table S5. For the ESM-2-35M model with LoRA, we train only 5.97M parameters (15.05\% of the total 39.67M), yet achieve performance improvements of up to 17.14\% on GB1. The ESM-2-150M model with LoRA requires training just 15.54M parameters (9.49\% of the total 163.71M) while delivering improvements of up to 9.65\% on Fold-superfamily. Even the largest ESM-2-650M model trains only 35.56M parameters (5.18\% of the total 685.80M) with LoRA. Beyond parameter efficiency, our analysis reveals favorable computational trade-offs. As shown in Table S5, while LoRA finetuning increases training time (2.6x for ESM-2-35M, 2.7x for ESM-2-150M), inference time remains minimal. This means that once trained, smaller LoRA-adapted models maintain fast inference speeds while delivering high performance.

Examining ESM-C models further validates these findings, with the ESM-C 300M achieving competitive performance to larger models across multiple tasks. Notably, ESM-C-300M with LoRA achieves 0.818 on Loc and 0.989 on Fold-family, showing that even mid-sized models can achieve strong results with efficient adaptation.

For resource-constrained applications such as online services, mobile devices, or research environments with limited GPU access, our findings provide clear guidance: applying LoRA to smaller PLMs offers an optimal balance between performance, computational efficiency, and deployment feasibility. The combination of reduced parameter count, minimal inference overhead, and superior task performance makes LoRA-adapted smaller models more attractive.

\subsection{Effects of Contact Information}

\begin{table}
\centering
\scriptsize
\caption{Performance comparison of different downstream prediction heads, SMH, MAH, and CM-MAH, using ESM-2-650M with LoRA finetuning. }
\label{tab:smh_mha_mha-c}
\begin{tabular}{c|c|c|c}
\toprule
\backslashbox{Dataset}{Head}    & SMH           & MAH                   & CM-MAH                \\ \midrule
EC               & 0.879±0.002                  & 0.884±0.003           & \textbf{0.886}±0.000  \\
GO-BP            & 0.460±0.003                  & \textbf{0.472}±0.003  & 0.463±0.011           \\
GO-MF            & 0.657±0.001                  & 0.673±0.003           & \textbf{0.674}±0.002  \\
GO-CC            & 0.538±0.001                  & \textbf{0.548}±0.007  & 0.534±0.015           \\
Fold-family      & 0.983±0.003                  & 0.982±0.006           & \textbf{0.988}±0.001  \\
Fold-fold        & 0.313±0.011                  & 0.334±0.007           & \textbf{0.339}±0.017  \\
Fold-superfamily & 0.699±0.006                  & 0.729±0.028           & \textbf{0.746}±0.006  \\
SS               & \textbf{0.826}±0.001         & 0.825±0.002           & 0.825±0.002           \\
Loc              & 0.820±0.001                  & 0.826±0.003           & \textbf{0.831}±0.003  \\
FLU              & \textbf{0.682}±0.003                  & \textbf{0.682}±0.001           & \textbf{0.682}±0.003  \\
HumanCell        & 0.704±0.005                  & \textbf{0.708}±0.003           & \textbf{0.708}±0.003  \\
GB1              & 0.953±0.000                  & 0.951±0.002           & \textbf{0.954}±0.003  \\
\bottomrule
\end{tabular}
\end{table}

To investigate whether contact information between protein amino acids can improve prediction accuracy, we systematically compared three downstream prediction heads using ESM-2-650M with LoRA finetuning across 10 tasks (12 subsets). Table~\ref{tab:smh_mha_mha-c} presents the comprehensive results comparing Simple MLP Head (SMH), Multi-head Attention Head (MAH), and Contact Map enhanced Multi-head Attention Head (CM-MAH). CM-MAH achieved the highest performance in most subsets, demonstrating the value of incorporating contact information. The most substantial improvements occurred in fold classification tasks: Fold-fold (+8.3\% over SMH, +1.5\% over MAH) and Fold-superfamily (+6.7\% over SMH, +2.3\% over MAH). 
MAH showed intermediate performance, outperforming SMH on most tasks but generally falling short of CM-MAH. This indicates that while multi-head attention provides benefits over simple linear layers, the integration of contact map information offers additional advantages.

\begin{table*}[t]
\centering
\scriptsize
\caption{
Performance comparison across different LoRA weight matrix configurations using ESM-2-35M with CM-MAH downstream head. Results show the impact of applying LoRA to various transformer components: query ($W_Q$), key ($W_K$), value ($W_V$) weight matrices, and their combinations, with ("+$W_d$") and without ("-") other dense layers. $\uparrow$: improvement with +$W_d$. }
\label{tab:loraqkv}
\resizebox{0.99\linewidth}{!}{%
\begin{tabular}{l|ccccccccccccc}
\toprule
\multirow{2}{*}{Module} & \multirow{2}{*}{Settings}  & \multirow{2}{*}{EC} & \multicolumn{3}{c}{GO} & \multicolumn{3}{c}{Fold} & \multirow{2}{*}{SS} & \multirow{2}{*}{Loc} & \multirow{2}{*}{FLU} & \multirow{2}{*}{HumanCell} & \multirow{2}{*}{GB1}  \\ \cmidrule{4-9}
 & & & BP & MF & CC & family & fold & superfamily & & & &   \\ \midrule
\multirow{4}{*}{$W_Q$} & \multirow{2}{*}{-} & 0.758  & 0.400  & 0.548  & 0.492  & 0.946  & 0.261  & 0.607  & 0.750  & 0.752  & 0.669  & 0.683  & 0.905  \\
&                   & ±0.001 & ±0.001 & ±0.000 & ±0.001 & ±0.006 & ±0.003 & ±0.012 & ±0.001 & ±0.001 & ±0.001 & ±0.001 & ±0.006  \\
                        & \multirow{2}{*}{+$W_{d}$}  & 0.786  & 0.413  & 0.571  & 0.497  & 0.963  & 0.274  & 0.606  & 0.750  & 0.757  & 0.681  & 0.687 & 0.938  \\
                        &                            & ±0.006 & ±0.001 & ±0.003 & ±0.004 & ±0.004 & ±0.008 & ±0.001 & ±0.001 & ±0.002 & ±0.002 & ±0.004 & ±0.001 \\
                        &                            & $\uparrow$ & $\uparrow$  & $\uparrow$  & $\uparrow$  & $\uparrow$  & $\uparrow$  & $\downarrow$  & $\simeq $  & $\uparrow$ & $\uparrow$  & $\uparrow$ & $\uparrow$  \\
                        \midrule
\multirow{4}{*}{$W_K$} & \multirow{2}{*}{-} & 0.766  & 0.404  &0.545  &0.492  & 0.943  & 0.271  & 0.603  & 0.749  & 0.754  & 0.672  & 0.682  & 0.901 \\
                        &                   & ±0.003 & ±0.001 &±0.001 &±0.002 & ±0.008 & ±0.006 & ±0.009 & ±0.001 & ±0.001 & ±0.001 & ±0.002 & ±0.003 \\
                        & \multirow{2}{*}{+$W_{d}$}  & 0.787  & 0.411  & 0.571  & 0.496  & 0.962  & 0.273  & 0.603 & 0.750  & 0.764  & 0.680  & 0.683 & 0.943 \\
                        &                            & ±0.006 & ±0.002 & ±0.001 & ±0.008 & ±0.001 & ±0.003 & ±0.007& ±0.000 & ±0.003 & ±0.000 & ±0.003 & ±0.002 \\ 
                       &                            & $\uparrow$ & $\uparrow$  & $\uparrow$  & $\uparrow$  & $\uparrow$  & $\uparrow$  & $\simeq$  & $\uparrow$ & $\uparrow$ & $\uparrow$  & $\uparrow$ & $\uparrow$  \\
                       \midrule
\multirow{4}{*}{$W_V$} & \multirow{2}{*}{-} &0.762  & 0.405  & 0.552  & 0.494  & 0.955  & 0.284  & \textbf{0.621}  & 0.750  & 0.758  & 0.669  & \textbf{0.695}  & 0.912  \\
                        &                   &±0.005 & ±0.002 & ±0.004 & ±0.003 & ±0.000 & ±0.004 & \textbf{±0.010} & ±0.001 & ±0.002 & ±0.002 & \textbf{±0.002} & ±0.004 \\
                        & \multirow{2}{*}{+$W_{d}$}  & 0.785 & 0.412  & 0.565  & 0.493  & 0.962  & 0.261  & 0.599  & 0.751  & 0.760  & 0.680  & 0.688 & 0.941   \\
                        &                            &±0.005 & ±0.002 & ±0.005 & ±0.007 & ±0.002 & ±0.015 & ±0.010 & ±0.001 & ±0.002 & ±0.001 & ±0.003 & ±0.001  \\  
                       &                            & $\uparrow$ & $\uparrow$  & $\uparrow$  & $\downarrow$  & $\uparrow$  & $\downarrow$  & $\downarrow$ & $\uparrow$   & $\uparrow$ & $\uparrow$  & $\downarrow$ & $\uparrow$  \\
                       \midrule
\multirow{4}{*}{\makecell{$W_Q$\\$W_K$}} & \multirow{2}{*}{-}  &0.763  & 0.403  & 0.553  & 0.499  & 0.957  & 0.275   & 0.610  & 0.751 & 0.750 & 0.677 & 0.678 & 0.918  \\
                                         &                     &±0.005 & ±0.002 & ±0.002 & ±0.003 & ±0.003 & ±0.012  & ±0.005 & ±0.001 & ±0.004 & ±0.002 & ±0.004 & ±0.003  \\
                            & \multirow{2}{*}{+$W_{d}$}        &0.788  & 0.410  & 0.572  & 0.500  & \textbf{0.963}  & 0.268  & 0.621  & 0.751 & 0.761 & 0.680 & 0.685 & 0.940  \\
                            &                                  &±0.004 & ±0.001 & ±0.004 & ±0.006 & \textbf{±0.001} & ±0.010 & ±0.006 & ±0.001 & ±0.000 & ±0.001 & ±0.003 & ±0.003  \\ 
                       &                            & $\uparrow$ & $\uparrow$  & $\uparrow$  & $\uparrow$  & $\uparrow$  & $\downarrow$  & $\uparrow$ & $\simeq$   & $\uparrow$ & $\uparrow$  & $\uparrow$ & $\uparrow$  \\
                       \midrule
\multirow{4}{*}{\makecell{$W_K$\\$W_V$}} & \multirow{2}{*}{-} &0.774  & 0.407  & 0.560  & 0.501  & 0.953  & 0.273   & 0.609  & 0.749  & 0.757  & 0.677 & 0.687  & 0.924   \\
                            &                                 &±0.003 & ±0.003 & ±0.004 & ±0.004 & ±0.008  & ±0.005 & ±0.007 & ±0.001 & ±0.004 & ±0.002 & ±0.004 & ±0.002  \\
                            & \multirow{2}{*}{+$W_{d}$}  &0.793  & 0.412  & 0.569  & 0.495  & 0.956 & 0.271  & 0.607  & 0.749  & 0.764  & 0.679  & 0.685  & 0.942  \\
                            &                            &±0.002 & ±0.004 & ±0.005 & ±0.008 & ±0.006 & ±0.010 & ±0.009 & ±0.002 & ±0.005 & ±0.002 & ±0.003 & ±0.003  \\    &                            & $\uparrow$ & $\uparrow$  & $\uparrow$  & $\downarrow$  & $\uparrow$  & $\downarrow$  & $\downarrow$ & $\simeq$   & $\uparrow$ & $\uparrow$  & $\downarrow$ & $\uparrow$  \\
                            \midrule
\multirow{4}{*}{\makecell{$W_Q$\\$W_V$}} & \multirow{2}{*}{-} &0.770  & 0.407  & 0.552  & 0.497  & 0.954  & 0.279   & 0.606  & 0.749  & 0.757  & 0.674 & 0.688  & 0.923   \\
                            &                                 &±0.005 & ±0.003 & ±0.000 & ±0.004 & ±0.006 & ±0.010  & ±0.011 & ±0.001 & ±0.004 & ±0.001 & ±0.003 & ±0.004 \\
                            & \multirow{2}{*}{+$W_{d}$}  &0.788  & 0.410  & 0.574  & 0.500  & 0.961 & 0.281  & 0.608  & 0.751  & 0.762  & 0.679  & 0.686  & 0.940 \\
                            &                            &±0.002 & ±0.002 & ±0.003 & ±0.004 & ±0.004 & ±0.005 & ±0.009 & ±0.002 & ±0.005 & ±0.002 & ±0.000 & ±0.002 \\                        &                            & $\uparrow$ & $\uparrow$  & $\uparrow$  & $\uparrow$  & $\uparrow$  & $\uparrow$  & $\uparrow$ & $\uparrow$   & $\uparrow$ & $\uparrow$  & $\downarrow$ & $\uparrow$  \\
                            \midrule
\multirow{4}{*}{\makecell{$W_Q$\\$W_K$\\$W_V$}} & \multirow{2}{*}{-} &0.775  & 0.408  & 0.557  & \textbf{0.505}  & 0.949  & \textbf{0.291}   & 0.606  & \textbf{0.751}  & \textbf{0.758}  & 0.677 & 0.684  & 0.929   \\
                            &                                        &±0.001 & ±0.002 & ±0.003 & \textbf{±0.005} & ±0.004 & \textbf{±0.010}  & ±0.007 & \textbf{±0.002} & \textbf{±0.003} & ±0.001 & ±0.003 & ±0.004 \\
                            & \multirow{2}{*}{+$W_{d}$}  & \textbf{0.793} & \textbf{0.411} & \textbf{0.573} & 0.494 & 0.953 & 0.276 & 0.596 & 0.750 & 0.756 & \textbf{0.681} & 0.689 & \textbf{0.941}  \\
                            &                           & \textbf{±0.001} & \textbf{±0.000} & \textbf{±0.001} & ±0.007 & ±0.006 & ±0.009 & ±0.010 & ±0.002 & ±0.003 & \textbf{±0.001} & ±0.005 & \textbf{±0.001}  \\
                       &                            & $\uparrow$ & $\uparrow$  & $\uparrow$  & $\downarrow$  & $\uparrow$  & $\downarrow$  & $\downarrow$ & $\downarrow$   & $\downarrow$ & $\uparrow$  & $\uparrow$ & $\uparrow$  \\
                       \bottomrule
\end{tabular}%
}
\end{table*}

When introducing the multi-head attention mechanism, MAH showed slight improvements in the EC and Fold tasks compared to the SMH, which stacks several linear layers. Specifically, MAH improved the EC task performance by 0.014 and the Fold task performance by 0.6. However, its performance on the regression tasks FLU and GB1 slightly declined.

CM-MAH exhibited more notable changes. In CM-MAH, we used contact information obtained by retraining the contact head of the ESM-2 model as the weight matrix for the multi-head attention. This adjustment led to significant performance improvements. For the EC task, CM-MAH improved by 0.027 compared to SMH and by 0.013 compared to MAH. In the Fold task, CM-MAH showed an improvement of 3.1 over SMH and 2.5 over MAH. In contrast, for the regression tasks, the performance of CM-MAH was almost the same as that of SMH and MAH, with a slight edge in the FLU task and matching SMH in the GB1 task.

For regression tasks, the improvements were relatively modest. CM-MAH matched or slightly improved performance on Loc, while maintaining comparable results on FLU, HumanCell, and GB1. This suggests that contact information provides greater benefits mostly for classification tasks that rely on structural relationships between residues, whereas regression tasks may depend more heavily on global sequence features.

These results suggest that training the contact head and incorporating contact information during finetuning enhances the model's ability to represent classification tasks effectively. However, this approach offers limited improvements for regression tasks. This difference could be due to the distinct feature requirements of classification and regression tasks. Classification tasks often depend on both local and global patterns within protein structures, which contact maps can provide effectively. In contrast, regression tasks are more focused on predicting continuous values and may rely more heavily on global sequence features rather than specific contact information. Although contact maps can enrich structural representation, this information may not directly contribute to performance enhancement in regression tasks. Future research could explore optimizing the use of contact maps or integrating them with other features to further improve model performance across various tasks.

\subsection{Analyzing Transformer Module Importance and Effect of Rank in LoRA Finetuning}

\begin{table*}
\begin{center}
\scriptsize
\caption{Impact of LoRA rank values on model performance using ESM-2-35M with CM-MAH downstream head. Results compare rank values $r \in \{1,2,4,8,16,32\}$ where higher ranks correspond to more trainable parameters.}
\label{tab:lorar}
\resizebox{0.9\linewidth}{!}{%
\begin{tabular}[0.7\linewidth]{c|cccccccccccc}
\toprule
\multirow{2}{*}{Rank}  & \multirow{2}{*}{EC} & \multicolumn{3}{c}{GO} & \multicolumn{3}{c}{Fold} & \multirow{2}{*}{SS} & \multirow{2}{*}{Loc} & \multirow{2}{*}{FLU} & \multirow{2}{*}{HumanCell} & \multirow{2}{*}{GB1} \\ \cmidrule{3-8}
 & & BP & MF & CC & family & fold & superfamily & & & &   \\ \midrule
\multirow{2}{*}{1} & 0.779  & 0.409  & 0.552  & 0.491  & 0.957  & 0.281  & 0.617  & 0.751  & 0.755  & 0.680  & 0.687  & 0.944 \\
                   & ±0.004 & ±0.002 & ±0.005 & ±0.006 & ±0.002 & ±0.003 & ±0.009 & ±0.002 & ±0.002 & ±0.001 & ±0.005 & ±0.002 \\ \midrule
\multirow{2}{*}{2} & 0.790  & 0.411  &0.562  &0.495  & 0.958  & 0.296  & 0.619  & 0.749  & 0.758  & 0.681  & 0.682  & 0.943 \\
                   & ±0.001 & ±0.002 &±0.004 &±0.003 & ±0.002 & ±0.001 & ±0.007 & ±0.000 & ±0.001 & ±0.003 & ±0.005 & ±0.001 \\ \midrule
\multirow{2}{*}{4} & 0.787  & 0.414  & 0.566  & 0.498  & 0.960  & 0.273  & 0.601  & 0.752  & 0.759  & 0.682  & 0.686  & 0.945  \\
                   & ±0.002 & ±0.002 & ±0.006 & ±0.005 & ±0.002 & ±0.009 & ±0.008 & ±0.001 & ±0.008 & ±0.002 & ±0.004 & ±0.001 \\ \midrule
\multirow{2}{*}{8} & 0.789 & 0.412 & 0.571 & 0.496 & 0.964 & 0.282  & 0.609 & 0.749 & 0.759 & 0.680 & 0.685 & 0.945  \\
                   & ±0.003 & ±0.003 & ±0.007 & ±0.005 & ±0.004 & ±0.004  & ±0.005 & ±0.001 & ±0.003 & ±0.001 & ±0.004 & ±0.001 \\ \midrule
\multirow{2}{*}{16} &0.791  & 0.412  & 0.575  & 0.494  & 0.967  & 0.279   & 0.599  & 0.751  & 0.762  & 0.680 & 0.684  & 0.941\\
                    &±0.002 & ±0.004 & ±0.005 & ±0.006 & ±0.002  & ±0.009 & ±0.010 & ±0.003 & ±0.002 & ±0.001 & ±0.006 & ±0.001 \\ \midrule
\multirow{2}{*}{32} & 0.793 & 0.411 & 0.573 & 0.494 & 0.953 & 0.276 & 0.596 & 0.750 & 0.756 & 0.681 & 0.689 & 0.941\\
                    & ±0.001 & ±0.000 & ±0.001 & ±0.007 & ±0.006 & ±0.009 & ±0.010 & ±0.002 & ±0.003 & ±0.001 & ±0.005 & ±0.001 \\
\bottomrule
\end{tabular}%
}
\end{center}
\end{table*}

\textbf{LoRA Weight Matrix Configuration Analysis}: We designed experiments to explore the effects of applying LoRA to various Transformer weight matrices (query, key, value, and other dense layers) across 10 downstream tasks. The results shown in Table~\ref{tab:loraqkv} reveal that applying LoRA to all transformer components in general shows the best performance in most tasks. The best-performing configurations vary slightly by task. For instance, Fold-family reaches peak performance (0.963) with $W_Q, W_d$ or $W_Q, W_K, W_d$ configurations. In these cases, the same peak performance, but with small SD values, is preferred. In addition, the inclusion of tuning LoRA in dense layers $W_d$ mostly shows enhanced performance across tasks. However, some tasks perform better without dense layers. Fold-fold achieves its highest accuracy (0.291) with $W_Q, W_K, W_V$ alone, and Fold-superfamily peaks (0.621) with $W_V$ alone.

\textbf{Rank Selection Analysis}: Our evaluation of ranks $r\in \{1,2,4,8,16,32\}$ (Table~\ref{tab:lorar}) suggests a comparable relationship between rank size and performance improvement. Higher ranks do not guarantee better results across tasks. Most tasks demonstrate relatively stable performance across different ranks, indicating that performance gains from increased rank are often marginal. These results suggest that lower ranks (r=1-8) can achieve comparable performance to higher ranks while requiring significantly fewer trainable parameters. This observation regarding rank selection partially aligns with~\cite{hu2022lora, sledzieski2024demo}, where optimal performance was reported at rank r=4 for protein-protein interaction prediction and r=8 for homooligomer symmetry prediction.

These results suggest that computational resources can be conserved by using fewer modules and lower ranks without sacrificing much accuracy. However, given that LoRA fine-tuning only brings a marginal increase in model parameters. Our results suggest that the use of LoRA in all components of the model, but with lower ranks, could potentially be the first attempt at initial screening for different protein tasks.

\subsection{Interpretability of LoRA Finetuning through Attention Analysis}

To understand how LoRA finetuning helps models identify important sequence features for classification tasks, we conducted a detailed analysis of attention patterns in ESM-2-650M models with and without LoRA adaptation. We selected samples from the Fold and Loc datasets that were incorrectly classified without applying LoRA finetuning but were correctly classified after LoRA finetuning. For each sample, we extracted and analyzed the attention matrices from all 33 transformer layers. For each layer, we extracted the raw attention matrices, averaged over the attention heads, and summed over columns. We then normalized these attention vectors to create comparable attention distributions. We stacked the normalized attention vectors from all layers to visualize the attention patterns across the entire model.

\begin{figure*}
\centering
\includegraphics[width=0.95\linewidth]{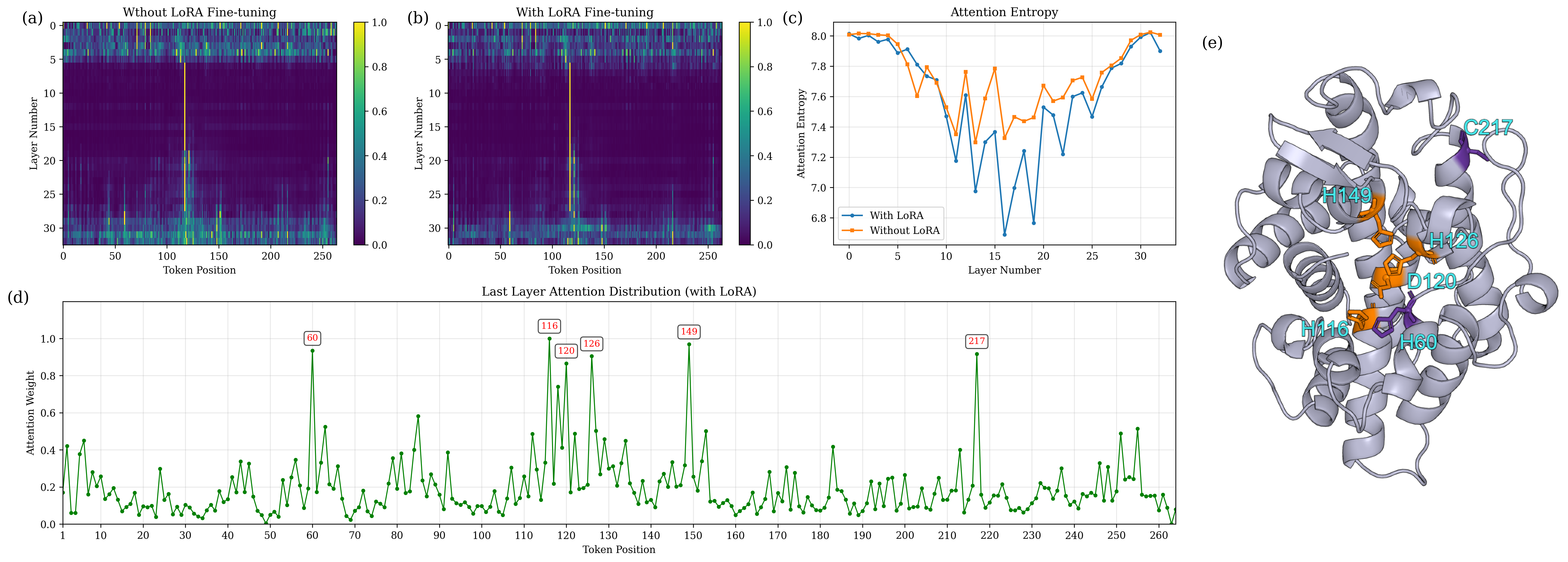}
\caption{Attention analysis of Nuclease P1 (PDB ID: 1AK0). (a) Layer-stacked attention heatmaps without LoRA finetuning. (b) Layer-stacked attention heatmaps with LoRA finetuning. The x-axis in (a) and (b) represents token positions in the protein sequence, and the y-axis represents transformer layers (a total of 33 layers). Higher attention weights are shown in brighter colors. (c) Comparison of per-layer attention entropy between models with and without LoRA fine-tuning. Lower entropy indicates a more focused attention distribution. (d) Attention weight distribution of the last transformer layer (with LoRA). The residue positions with high attention are labeled with a red number. (e) 3D structure of Nuclease P1. Known substrate binding site residues (H116, D120, H126, and H149) are highlighted in orange. Other high-attention residues (H60 and C217) are highlighted in purple.}
\label{fig:1ak0_att}
\end{figure*}

Figure~\ref{fig:1ak0_att} (a) and (b) show the layer-stacked attention heatmaps for Nuclease P1 (PDB ID: 1AK0) selected from the Fold dataset. The heatmaps reveal that the LoRA finetuned model exhibits more focused attention patterns, particularly in functionally important regions. For example, the LoRA finetuned model shows enhanced attention at residues H116, D120, H126, and H149 (Figure~\ref{fig:1ak0_att} (d-e)), which coincides with the known substrate binding sites of Nuclease P1 according to \cite{romier19981ak0}. This alignment between model attention and known functional regions suggests that LoRA finetuning helps the model identify biologically relevant sequence features.

To quantify these differences, we calculated the attention entropy for each layer, with results shown in Figure~\ref{fig:1ak0_att} (c). Attention entropy measures how evenly distributed the attention is across residues \textemdash lower entropy indicates more focused attention on specific residues. LoRA finetuning consistently reduces attention entropy across most layers, with particularly pronounced differences in middle-to-deep layers (12-20).

We extended this analysis to 10 additional samples from the Fold (Figure S11-S15) and Loc (Figure S16-Figure S20) datasets. Across this larger sample, we observed consistent patterns:
1. LoRA finetuned models showed a consistent reduction in attention entropy compared to non-LoRA-finetuned models.
2. The middle layers (12-20) of the transformer consistently showed the greatest adaptation to attention patterns.

These findings indicate that LoRA adaptation enables the model to learn more selective attention patterns that focus on task-relevant sequence features rather than distributing attention broadly across the sequence. Moreover, LoRA adaptation primarily affects higher-level feature extraction rather than low-level sequence processing, which is consistent with the theory that lower layers of protein language models capture more general sequence patterns, while deeper layers represent more task-specific features.

\subsection{Sensitivity and Specificity Analysis of Classification Performance}

\begin{table*}[th]
\centering
\scriptsize
\caption{The results of the compared methods using sequence only or additional data features in all datasets. SeqProFT model version is the ESM-2-650M with CM-MAH and LoRA finetune. Values of the compared models are cited from original reports. Black bold text: the best performance among non-sequence-only models. \textcolor{purple}{Purple bold text: the best performance among sequence-only models.}}
\label{tab:vssota1}
\begin{tabular}{c|c|cccccccccccc}
\toprule
\multirow{2}{*}{Methods} & \multirow{2}{*}{\parbox{0.5cm}{\centering Seq. Only}} & \multirow{2}{*}{EC} & \multicolumn{3}{c}{GO} & \multicolumn{3}{c}{Fold} & \multirow{2}{*}{SS} & \multirow{2}{*}{Loc} & \multirow{2}{*}{FLU} & \multirow{2}{*}{HumanCell} & \multirow{2}{*}{GB1}\\ \cmidrule{4-9}
 & & & BP & MF & CC & family & fold & superfamily & & & & &     \\
\midrule
DeepFRI\cite{gligorijevi2021deepfri} & x & 0.631  & 0.399 & 0.465 & 0.460 & 0.732 & 0.153 & 0.206 & - & -& -& -& -\\
GearNet\cite{zhang2023gearnet} & x & 0.730  & 0.356 & 0.503 & 0.414 &0.953&0.284&0.426& - & -& -& -& -\\
GearNet-MC\cite{zhang2023gearnet} & x  & 0.874  & 0.490 & 0.654 & 0.488&\textbf{0.999}&\textbf{0.541}&\textbf{0.805}& - & -& -& -& - \\
ProtST-ESM-2\cite{xu2023protst} & x & 0.878  & 0.482 & 0.668 & 0.487 & - & -& -& -& 0.802& 0.682 & 0.682& -\\
ESM-1B-GearNet\cite{zhang2023enhancing} & x  & 0.883  & 0.491 & 0.677 & 0.501 & - & -& -& -& -& - & -& -\\
ESM-2-GearNet\cite{zhang2023systematic} & x  & \textbf{0.897}  & 0.514 & \textbf{0.683} &  \textbf{0.505} & - & -& -& -& -& - & -& -\\
ESM-2-RS\cite{zhang2024esms} & x &  0.817 & \textbf{0.519} & 0.678 & 0.485& - & -& -& -& -& - & -& - \\
ProteinINR\cite{lee2024proteininr}  & x  & 0.896  & 0.518 & \textbf{0.683} & 0.504 & - & -& -& -& -& - & -& -\\
\midrule
CNN\cite{shanehsazzadeh2020cnn} & \checkmark & 0.545  & 0.244 & 0.354 & 0.287 &0.534&0.113&0.134 & - & -& -& -& -\\
Ankh \cite{elnaggar2023ankh} & \checkmark & - & -& -& -& -& -&0.611&0.838&\textbf{\textcolor{purple}{0.832}}&0.620&-&0.840 \\
ESM-1B \cite{xu2023protst} & \checkmark& 0.869  & 0.452 & 0.659 & 0.477& - & -& -& -&0.781&0.679&0.699&- \\
ESM-2 \cite{xu2023protst}  & \checkmark & 0.874  & \textcolor{purple}{\textbf{0.472}} & 0.662 & 0.472 & - & -& -& -&0.787&0.677&0.672&- \\
ESM-2 \cite{elnaggar2023ankh}  & \checkmark & - & -& -& -& -& -&0.563&0.823&0.818&0.480&-&0.820\\
ESM-2 \cite{schmirler2023finetunetum}   & \checkmark & - & -& -& -& -& -&-&\textbf{\textcolor{purple}{0.855}}&0.638&\textbf{\textcolor{purple}{0.688}}&-&- \\
\midrule
SeqProFT  & \checkmark & \textcolor{purple}{\textbf{0.886}} & 0.463 & \textcolor{purple}{\textbf{0.674}} & \textbf{\textcolor{purple}{0.534}} &\textcolor{purple}{\textbf{0.988}}&\textcolor{purple}{\textbf{0.339}}&\textcolor{purple}{\textbf{0.746}}&0.825&0.831&0.682&\textbf{\textcolor{purple}{0.708}}&\textbf{\textcolor{purple}{0.954}}\\

\bottomrule
\end{tabular}%
\end{table*}

To provide a comprehensive evaluation of the quality of the model prediction in different classes, we calculated the macro-averaged sensitivity and specificity for all classification tasks.
\begin{itemize}[nosep]
    \item Macro-Sensitivity$=\frac{1}{C}\times \sum_{c=1}^C\frac{TP_c}{TP_c+FN_c}$: the proportion of actual positive cases correctly identified,
    \item Macro-Specificity$=\frac{1}{C}\times \sum_{c=1}^C\frac{TN_c}{TN_c+FP_c}$: the proportion of actual negative cases correctly identified,
\end{itemize}
where C is the total number of classes, and $TP_c$, $FN_c$, $TN_c$, $FP_c$ are the true positives, false negatives, true negatives, and false positives for class c, respectively. Table S6 presents the results for SeqProFT using ESM-2-650M with CM-MAH, comparing the performance with and without LoRA finetuning.

\textbf{Sensitivity Analysis}: LoRA finetuning significantly improves sensitivity across most tasks, with the largest gains observed in Fold-superfamily (+24.67\%), GO-MF (+22.02\%), and Fold-fold (+18.54\%). These improvements indicate that LoRA enables better identification of positive cases across different protein classes. Notably, EC classification shows an 11.66\% improvement, demonstrating enhanced enzyme function prediction capabilities. The only exception is GO-CC, which shows a modest decrease (-4.33\%), suggesting task-specific adaptation effects.

\textbf{Specificity Analysis}: All tasks maintain excellent specificity (>0.976), with LoRA finetuning providing consistent but modest improvements (+0.01\% to +0.43\%). The high baseline specificity values indicate that both models effectively minimize false positive predictions across protein classes. The smaller improvement margins reflect the already strong performance in correctly identifying negative cases.

The results reveal different adaptation patterns in tasks. For multi-label tasks (EC, GO), LoRA primarily enhances sensitivity while maintaining high specificity. The sensitivity improvements for Fold tasks are significant, given the large number of classes (1,195 folds), while improvements are more balanced for SS (3 classes) and Loc (10 classes). These findings suggest that LoRA finetuning is particularly effective for improving recall in complex protein classification scenarios where positive identification is challenging.

\subsection{Comparison with the State-Of-The-Art Methods}

Table~\ref{tab:vssota1} shows the comparison of our model with other state-of-the-art models. In this analysis, our SeqProFT model is with ESM-2-650M and CM-MAH finetuned using LoRA (r=32, alpha=32). The models in \cite{gligorijevi2021deepfri}, \cite{zhang2023gearnet}, \cite{zhang2023enhancing}, \cite{zhang2023systematic}, \cite{zhang2024esms}, and \cite{lee2024proteininr} are not sequence-only and utilize additional protein structure information, 
while \cite{xu2023protst} incorporates biomedical text data as a new modality. All other methods use only protein sequence data for model training. It is important to note that ESM-2 (from~\cite{schmirler2023finetunetum, xu2023protst}) is fully finetuned. Our approach shows similar or better performance compared to other sequence-only models. In some tasks, SeqProFT shows the best performance, compared to those models with additional modalities. For GO-CC, SeqProFT outperforms all other models with structure information. For HumanCell, SeqProFT shows better performance than the model with additional text modality~\cite{xu2023protst}. For GB1, SeqProFT is significantly better than Ankh\cite{elnaggar2023ankh}, which has 1.5 billion parameters.

While our method does not fully match the performance of models using additional modalities, it demonstrates reasonable comparability. This indicates that SeqProFT can be a valuable option for predicting protein properties in datasets lacking structural information and only using sequence as input. Furthermore, our systematic evaluation (Table~\ref{tab:mha-c_loraVSnolora}) shows consistent performance improvements with LoRA finetuning, and the parameter efficiency analysis demonstrates that these gains are achieved while training only 5.18\%-15.05\% of total model parameters. Robust performance in various tasks suggests that SeqProFT effectively balances computational efficiency and predictive accuracy for the prediction of sequence-only protein properties.

\section{Conclusion and Discussion}
\label{sec:Conclusion}

This study proposes SeqProFT, a parameter-efficient framework for the prediction of sequence-only protein properties that combines LoRA finetuning with multihead attention mechanisms and contact map integration. Our systematic evaluation on 10 diverse downstream tasks and 12 datasets demonstrates that SeqProFT shows excellent performance on both classification and regression tasks. Our experiments reveal that different parameter configurations of LoRA give robust model performance for different tasks, highlighting the importance of tailored finetuning strategies. Notably, increasing model complexity does not always yield improvements, where we observed that larger PLMs are not consistently superior. Smaller models (35M-300M parameters) with LoRA adaptation can match or exceed the performance of models 4-5 times larger, while requiring only a fraction of the computational resources. This finding provides benefits in the implementation of advanced protein analysis in resource-constrained environments. For small laboratories with limited resources or online platforms, LoRA finetuning of smaller PLMs offers an effective and practical alternative to large model deployment, providing an optimal balance between performance and computational efficiency. Furthermore, our attention analysis suggests that LoRA tuning helps identify biologically critical regions within protein sequences. This could provide valuable new insights for drug discovery, for example, targeting specific mutations and understanding disease mechanisms.

\ifCLASSOPTIONcaptionsoff
  \newpage
\fi
\bibliographystyle{IEEEtran}
\bibliography{mybibfile}

\ifCLASSOPTIONcaptionsoff
\vspace{-15 mm}

\fi

\end{document}